\documentclass{article}

\usepackage{arxiv}

\usepackage[utf8]{inputenc} 
\usepackage[T1]{fontenc}    
\usepackage{hyperref}       
\usepackage{url}            
\usepackage{booktabs}       
\usepackage{amsfonts}       
\usepackage{nicefrac}       
\usepackage{microtype}      
\usepackage{lipsum}		
\usepackage{graphicx}
\usepackage{natbib}
\usepackage{doi}

\usepackage{amsmath}
\usepackage{subcaption}
\usepackage{array}
\usepackage{booktabs}
\usepackage{comment}

\title{Learning UI Navigation through Demonstrations composed of Macro Actions}

\author{ Wei Li\\
	Google Research\\
	New York, NY 10011 \\
}

\date{}

\renewcommand{\headeright}{ }
\renewcommand{\undertitle}{ }
\renewcommand{\shorttitle}{Learning UI Navigation}

\hypersetup{
pdftitle={Learning UI Navigation through Demonstrations composed of Macro Actions},
pdfsubject={q-bio.NC, q-bio.QM},
pdfauthor={Wei Li},
pdfkeywords={UI navigation, imitation learning},
}

\begin{document}
\maketitle

\begin{abstract}
We have developed a framework to reliably build agents capable of UI navigation. The state space is simplified from raw-pixels to a set of UI elements extracted from screen understanding, such as OCR and icon detection. The action space is restricted to the UI elements plus a few global actions. Actions can be customized for tasks and each action is a sequence of basic operations conditioned on status checks. With such a design, we are able to train DQfD and BC agents with a small number of demonstration episodes. We propose demo  augmentation that significantly reduces the required number of human demonstrations. We made a customization of DQfD to allow demos collected on screenshots to facilitate the demo coverage of rare cases. Demos are only collected for the failed cases during the evaluation of the previous version of the agent. With 10s of iterations looping over evaluation, demo collection, and training, the agent reaches a 98.7\% success rate on the search task in an environment of 80+ apps and websites where initial states and viewing parameters are randomized.
\end{abstract}


\section{Introduction}
The goal is to build agents capable of human-like actions by navigating through graphical or textual user interfaces (UI), such as web pages and mobile phones. Applications of such intelligence include but are not limited to automated task completion, such as installing and updating software, tutorial or assistance to humans on how to operate devices, and with the help of natural language processing, following user commands or following user manuals. Typically, a task is associated with an utterance that specifies what sub-task to complete and optionally additional arguments that are needed for the sub-task. For example, in a follow-user-command task, a utterance could be “search for something here”, and the expected behavior for an agent is to locate the search bar on the screen and input ‘something’ before pressing the enter key. If a search bar is not visible on the current screen, the agent should perform proper action to reveal it.

The input of these agents are preferably raw pixels that are presented to humans as modern UI systems aimed at smooth user journeys by providing sufficient information of the current state and are intuitive to perform desired actions. Relying on pixels avoids the complexity of implementing automation interfaces in each software application as any human-facing app naturally comes with a UI.

There are existing automations that control applications or the whole device through programs that are also referred to as macros. Obviously, the creation of such macros requires coding by software engineers and their logics depend on the heuristics generalized by humans. Typically, such heuristics are structured as a sequence of if statements, each handling a group of scenarios. 
\begin{itemize}
\item If in scenario s1, performs action a1 on element e1;
\item else if in scenario s2, performs action a2 on element e2;
\item else if in scenario s3, performs action a3 on element e3;
\item{...}
\end{itemize}

An example scenario could be: a back arrow appears at the top-left corner of the screen; an example action could be a click; an element could be a button labeled with a back arrow.

When a failed scenario is encountered, it requires further software engineering to either adjust existing if-statements or add new branches. Both the creation and the maintenance of these automation macros are expensive as they involve software engineers.

In this paper, we describe a system that creates UI navigation agents with the power of neural networks that learn from human demonstrations. The key differentiation from previous work include:
\begin{itemize}
\item Instead of directly working on raw pixels, the input to the neural network is a collection of UI elements that are extracted from screen understanding modules, such as OCR, icon recognition, and image detection. This significantly reduces the state space.
\item Each action issued from the agent is executed as a program/macro that is composed of a sequence of basic or micro actions with branches controlled by status checks. From the agent’s point of view, regardless of the complexity of a macro action, it is atomic. In other words, while a macro action is pending, the agent simply waits for it to finish either successfully or with a failure. This greatly simplifies the action space. Transitional screens during the execution of a macro are not fed to the agent, hence does not contribute to the state space. Reduced state space and action space result in simplified neural networks that require fewer training samples and less training time.
\item We designed a neural network architecture that is analogous to the sequence of the if-statements in a heuristic agent. The neural nets can be trained by either deep Q-learning from demonstration (DQfD)\cite{dqfd} or behavior cloning (BC)\cite{pomerleau:alvinn,osa:2018} algorithms. Agents evolve through iterations over agent evaluation, demo collection, and neural network training. Demonstrations are collected in an error-driven fashion where failed scenarios during evaluation are recorded and reproduced. Only new demos that introduce different behavior in the current iteration of the agent are added to the training samples for the next iteration of the agent.
\item Demonstrations are collected using the same set of macro actions of the agent.
\item Human demonstrations are augmented to synthesize more demonstrations utilizing the information specific to demos, such as which UI elements are operated and prior knowledge on which UI elements are more important for UI navigation.
\item In addition to full-episode demonstrations, we also provide the option of using screenshot demos that facilitate the coverage of rare cases in training samples.
\end{itemize}

This paper focuses on UI navigation on Android, but our techniques apply to other UI systems as well, such as Web navigation using a DOM tree, and UI navigation through the accessibility api.

In the remainder of this paper, we will provide more technical details, followed by experimental results on a couple of reinforcement learning tasks on Android that we built. Finally, we conclude the paper.

\section{Related Work}
The works on MiniWoB\cite{miniwob}, where reinforcement learning is applied to Web-based applications, has the most resemblance to our work. \citet{wge} constrains agent's exploration according to demonstration to accelerate the discovery of rewards. \citet{learning_web_nav, gminiwob} use curriculum learning to decompose a difficult task into multiple easier sub-tasks. In contrast, our approach is essentially supervised learning, as our agents can be trained purely on demos, although our DQfD agent can be further tuned by interacting with the environment. We design the state space and action space to be based on UI elements. The abstract architecture and the demo augmentation make our neural nets simple to train that only require a few hundreds of human demonstrations to reach near perfect success rates. The usage of screen representation based on UI elements and macro action makes our architecture compatible to navigating Web and accessibility tree so that we leverage successful algorithms developed on MiniWoB.

\citet{mapping_instruction_to_action, reading_between_the_lines} applied Reinforcement Learning (RL) in two domains: Windows trouble shooting guides and game tutorials. 
\citet{nl_to_ui_action} map natural language instruction to mobile UI action sequences. Their work focuses on natural language processing. Both of them map spans in multi-step instructions to UI elements. Typically, each UI element to be operated is explicitly mentioned in the instruction. In comparison, we focus on how an single utterance is executed reliably on various applications with high success rate so that the technique can be productionized. The operation needed is not necessarily  included in the utterance. For example, for the utterance 'search for something here', an agent may need multiple clicks on various UI elements to dismiss popup ads, to switch pages, and to reveal the search bar.  

\section{Technical Details}
\subsection{Screen representation}
\begin{figure}
\centering
\includegraphics[width=0.5\columnwidth]{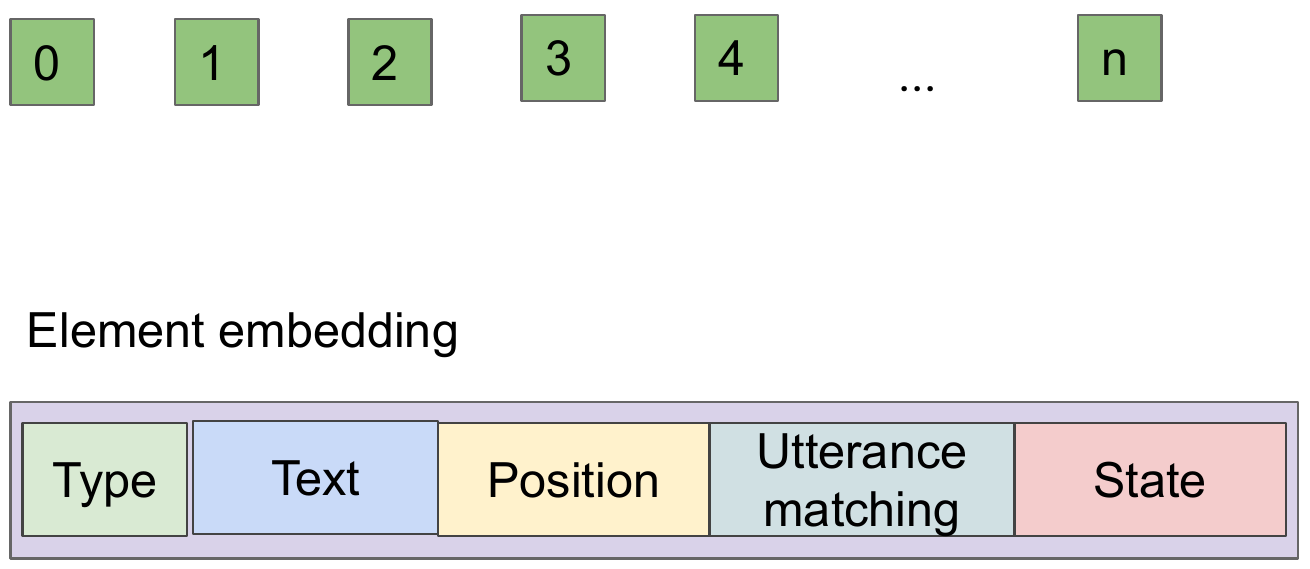} 
\caption{A screen is represented as a list of UI elements, of which each is encoded by concatenating various feature vectors of the associated attributes.}.
\label{fig_screen_representation}
\end{figure}

In our system, each screen is represented as a set of UI elements. As shown in Figure \ref{fig_screen_representation}, the numbered green boxes represent UI elements. Each UI element is encoded by concatenating the feature vectors of various attributes associated with the element, including its type, description text, screen position, etc. Some UI elements, such as radio buttons and checkboxes also have states.

The screen representation is an abstract structure and does not specify how it is obtained. For Android, it can be assembled from the output of screen understanding components that take raw pixels as input, such as OCR, icon recognizer, and image detector. For web navigation, it can be simplified from DOM trees. Besides, accessibility tree, if available, may be used to fully or partially populate the data of the screen representations.

\subsection{Action Space}
\begin{figure}
\centering
\includegraphics[width=0.7\columnwidth]{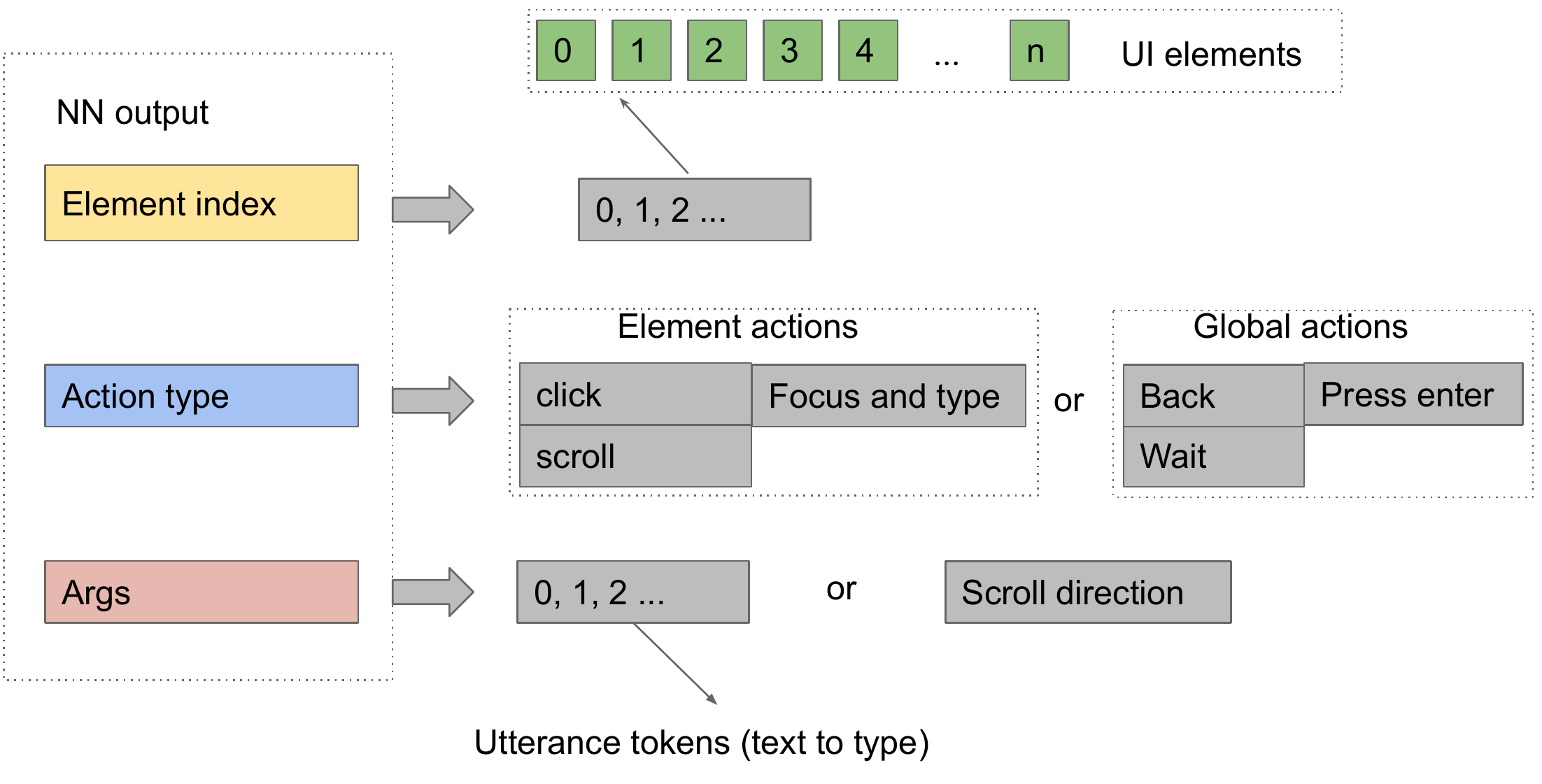} 
\caption{Action space}.
\label{fig_action_space}
\end{figure}

As shown in Figure \ref{fig_action_space}, we support two groups of actions:
\begin{itemize}
\item Element action: that is performed on an individual element. Examples are: click, focus\_and\_type (set focus to an UI element and type into it), scroll.
\item Global action: that is not pertinent to any element. Examples are: wait, back, press enter.
\end{itemize}
The output of an agent defines what action will be performed:
\begin{itemize}
\item Element index: identifies which element the action will be applied to.
\item Action type: specifies the type of action. The types include both element actions and global actions.
\item Action args: some types of actions require additional arguments. For example, for a scroll action, its direction is needed. Unless there is only one option, a focus\_and\_type needs an index into an array of phrases that can be entered.
\end{itemize}

\subsection{Neural network architecture}
\begin{figure}
\centering
\includegraphics[width=0.7\columnwidth]{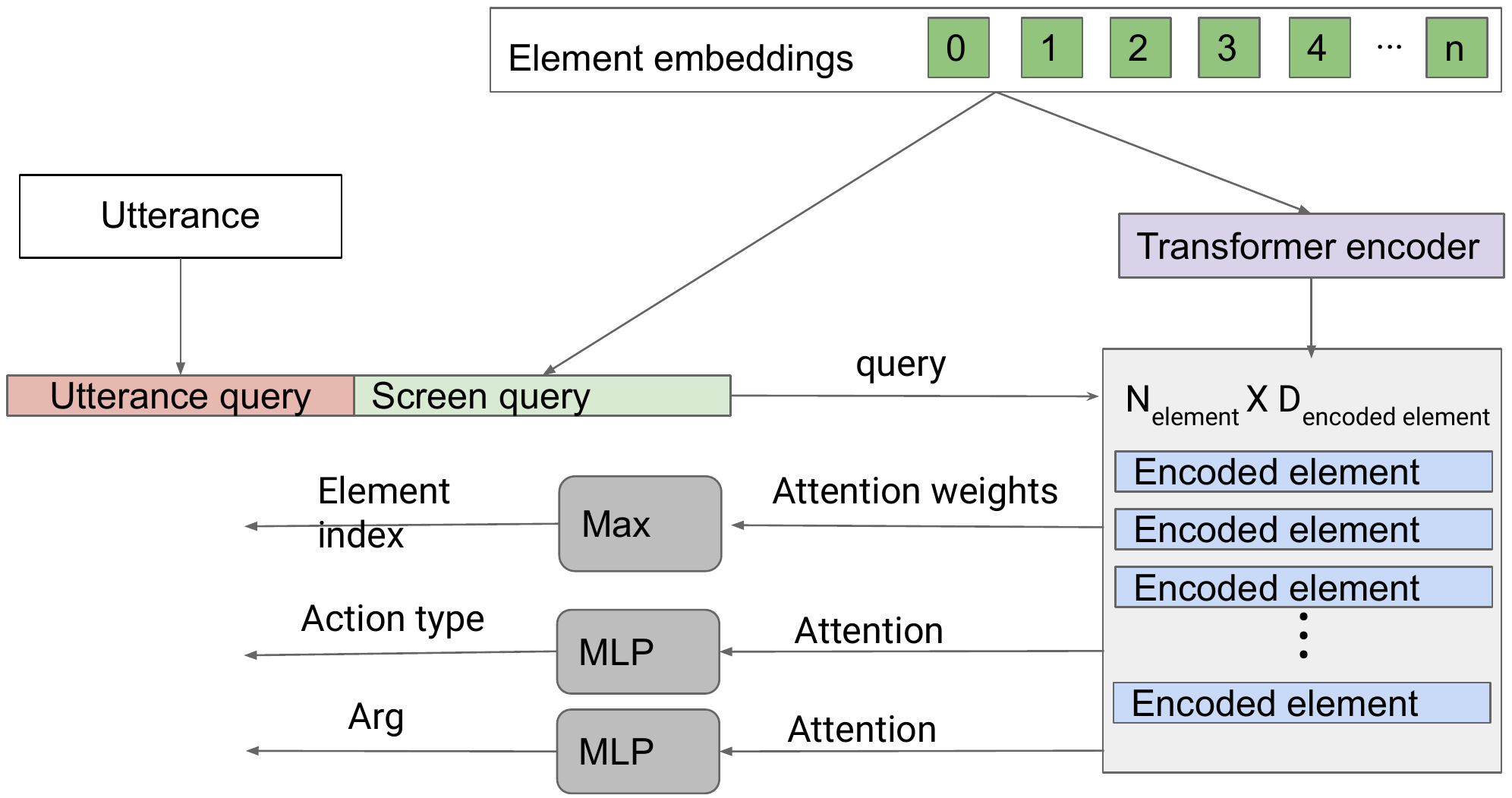} 
\caption{Neural network architecture}.
\label{fig_nn}
\end{figure}

Figure \ref{fig_nn} illustrates the neural network architecture for UI navigation. A screen representation of a set of element embeddings is encoded by a transformer encoder, whose output has the same number of entries as the number of UI elements. Intuitively, each entry is the encoding of the corresponding UI element plus its attention over all the UI elements on screen, including itself.

In order to select which UI element to perform the next action, a query vector is created from the concatenation of the utterance and the aggregate of the screen representation (such as the sum, average, or max of the element embeddings) for an attention mechanism over the encoded elements. The max of the attention weights is used to pick the element index and the attention is passed through two independent MLP modules to produce action type, and action argument.

There are multiple output heads from the neural net, while for training to apply gradient descent, the loss function should produce a single scalar variable. A natural choice is to set the overall loss as the sum of the losses of all the output heads. The multiple output heads are not independent of each other. For example, for a global action, the element index is ignored, and the loss contributed from the head of the element index should not affect the weights of the neural net. We use the following formula to compute the loss:

\begin{equation} \label{eq:1}
\begin{split}
&loss=loss_{action \, type} +loss_{element \, index} * mask_{element \, index}
    + loss_{action \, arg} * mask_{action \, arg}\\
&mask_{element \, index}= \begin{cases} \begin{matrix} 1, \text{if element action} \\ 0, \text{otherwise} \end{matrix} \end{cases}\\
&mask_{action \, arg} = \begin{cases} \begin{matrix} 1, \text{if action requires argument} \\ 0, \text{otherwise} \end{matrix} \end{cases} \\
\end{split}
\end{equation}

We found that masking out unneeded losses improves the training accuracy (Please see Table \ref{table3}). 

The neural net in figure \ref{fig_nn} can be considered as a scoring system. For any given screenshot, all the UI elements are scored, and the one with the highest score is selected for possible action. Note that due to attention in the transformer encoder, for any UI element, its relationship to all the UI elements on screen are encoded. This makes it possible for the neural net to include all the relevant combinations of UI elements that it encounters during training as states. The neural net maintains a ranking list of all these states, while each state maps to a UI element, an action type, and an action argument. The ranking list is equivalent to the ordered if-statements of the heuristic agent. We can imagine that the neural network traverses the ranking list at inference time in descending order. If a state has a match on the current screen, the corresponding UI element,  action type, and action argument are selected.

\subsection{Macro actions}

In our system, an agent action is a composite of a sequence of low level actions. We call the low level actions micro actions, and refer to composite actions as macro actions. Although viewed as an atomic step, a macro action is controlled by a program, also called a macro, and can involve arbitrary logic. Using macro actions has the following advantages:

\begin{itemize}
    \item Fewer steps to complete a task than using micro actions as each macro action contains multiple micro actions. This reduces the complexity that an agent faces.
    \item During the execution of an atomic macro action, changes to the screen are not visible to the agent, and do not contribute to the state space. In particular, each macro action is designed to skip transitional screens and finishes when the screen becomes stable or a timeout is reached when dealing with dynamic screens such as playing a video.
    \item A macro can depend on other macro for lower level actions. In other words, the organization of macros can be hierarchical.
    \item Using abstract macro actions makes it possible to build cross-platform agents. The definition of focus\_and\_type action is actually borrowed from MiniWoB\cite{miniwob,wge}.
\end{itemize}

\begin{figure}
    \centering
    \subfloat[]{\includegraphics[width=0.25\linewidth]{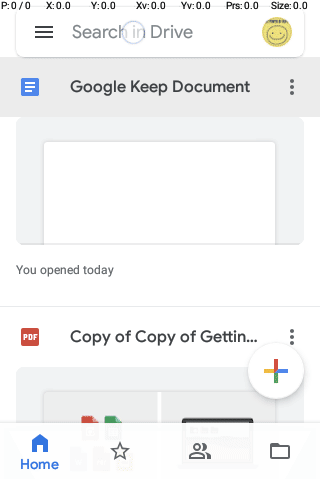}}
    \subfloat[]{\includegraphics[width=0.25\linewidth]{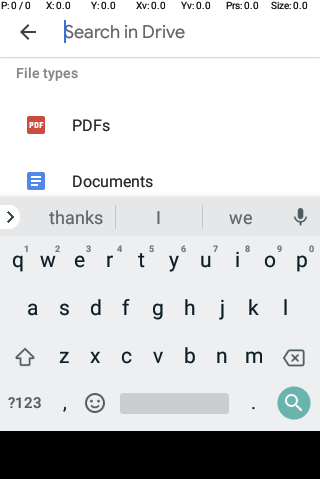}}
    \subfloat[]{\includegraphics[width=0.25\linewidth]{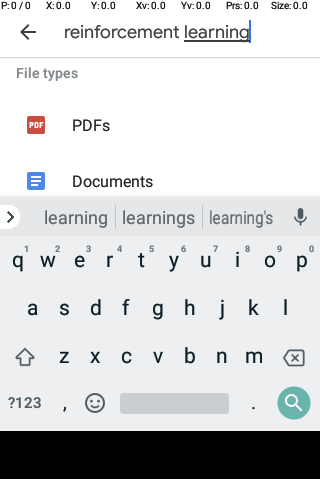}}
    \subfloat[]{\includegraphics[width=0.25\linewidth]{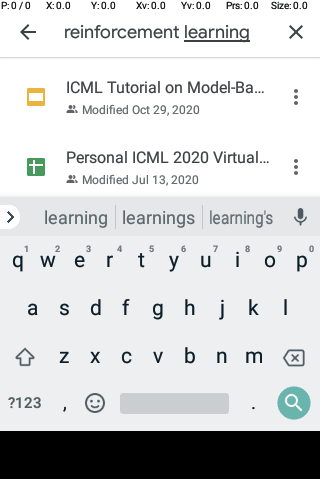}}
    \caption{An example macro action: focus\_and\_type. (a) click to set focus. (b) wait for blinking cursor. (c) type text. (d) press enter and wait screen changes.}
    \label{fig:focus_and_type}
\end{figure}

The image sequence in Figure \ref{fig:focus_and_type} shows the four steps of the focus\_and\_type action on Android that enters text in the field labelled “Search in Drive”:
\begin{enumerate}
    \item click the field to obtain focus (\ref{fig:focus_and_type}(a)).
    \item wait for the blinking cursor to appear (\ref{fig:focus_and_type}(b)).
    \item input text (\ref{fig:focus_and_type}(c)).
    \item press enter and wait for the screen to change (\ref{fig:focus_and_type}(d)).
\end{enumerate}

We have designed a blinking cursor detector. If the changed pixels in contiguous screenshots form a vertical bar, a candidate cursor is found. If a candidate repeatedly appears at the same location, a blinking cursor is found. For efficiency, the search region is restricted to the text input field, and vertical bars of unreasonable width or height are filtered out.

When an agent issues a click action, the following steps take place:

\begin{enumerate}
    \item trigger a click on a device (a pair of touch down and up actions for touch devices).
    \item wait for the expected screen change.
\end{enumerate}

Generally, an expected screen change after clicking a button is that a large percentage of pixels change. Clicking a radio button or checkbox does not comply with this assumption and can rely on a dedicated state recognizer.

When using screen understanding to generate screen representations, each macro action finishes with one of the following states:
\begin{itemize}
    \item SUCCESS
    \item FAILURE
    \item CANCELATION
\end{itemize}

The meaning of the first two states are obvious. The state of CANCELATION is designed to handle the delay of screen understanding. When an agent is about to perform an action on a UI element, it is possible that the UI element has changed (moved or disappeared) in the latest screen. If it is true, the action should be cancelled and the agent should pick a new action based on the latest screen. To detect whether a UI element has changed, the screenshot from which the element is recognized is cached and is compared with the latest screenshot within the bounding rectangle of the element. The latest screenshot is fed for comparison without the computation of screen understanding, hence much incurs much smaller delays.

When screen representation comes from a DOM tree or accessibility tree, an action always succeeds. Besides, after a reference to an element is obtained, action can be applied no matter where the element moves. The blinking cursor detection is also unnecessary, because the type of any text input field is known with no risk of mistake as it is part of the attributes of the corresponding tree nodes.

\subsection{Evaluation, demo collection, and training}

\begin{figure}
\centering
\includegraphics[width=0.7\columnwidth]{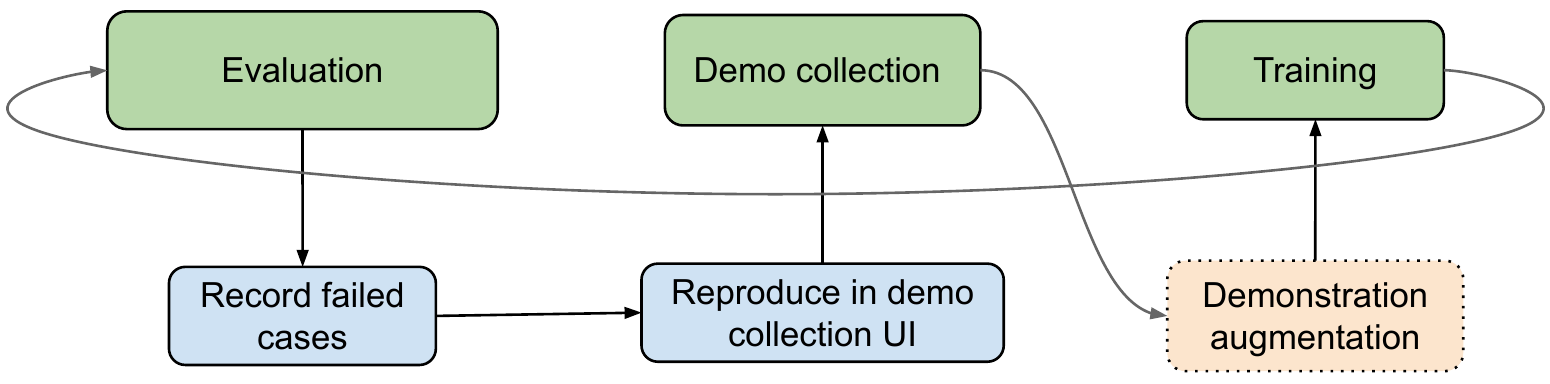} 
\caption{The loop for building UI navigation agents.}.
\label{fig:agent_building_loop}
\end{figure}

As shown in Figure \ref{fig:agent_building_loop}, a UI navigation agent is built through iterations looping over evaluation, demo collection, and training. The evaluation of a trained agent is done automatically if an RL environment is available where episodes of the task are terminated with a reward if successful or no reward if reached max number of steps. The process is performed manually where a human rater makes the success or failure decisions while watching an agent completing tasks. During an evaluation session of an agent, the configurations of all the failed cases are recorded. Then those failed cases are reproduced in the GUI of the demo collection and new demonstrations from humans are added to the pool of training samples if those demonstrations have different behavior than the current version of the agent. In short, demo collection is error-driven. Next, a new version of the agent is trained with all the demonstrations, and the steps repeat.

\begin{figure}
    \centering
    \subfloat[]{\includegraphics[width=0.5\linewidth]{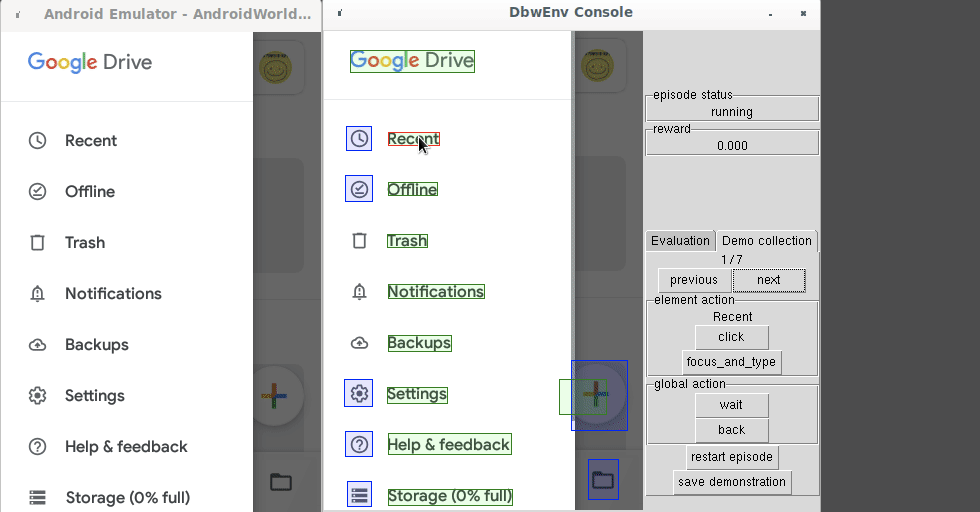}}
    \subfloat[]{\includegraphics[width=0.5\linewidth]{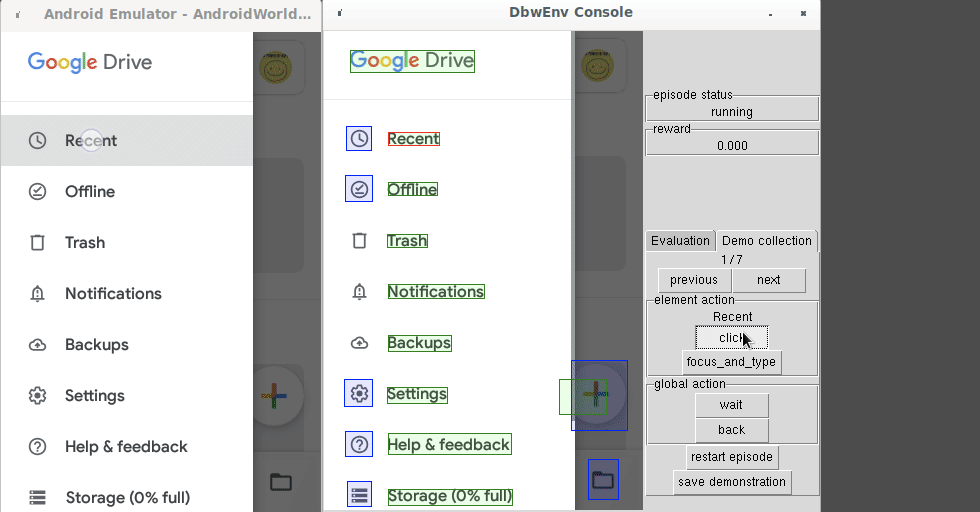}}\\
    \subfloat[]{\includegraphics[width=0.5\linewidth]{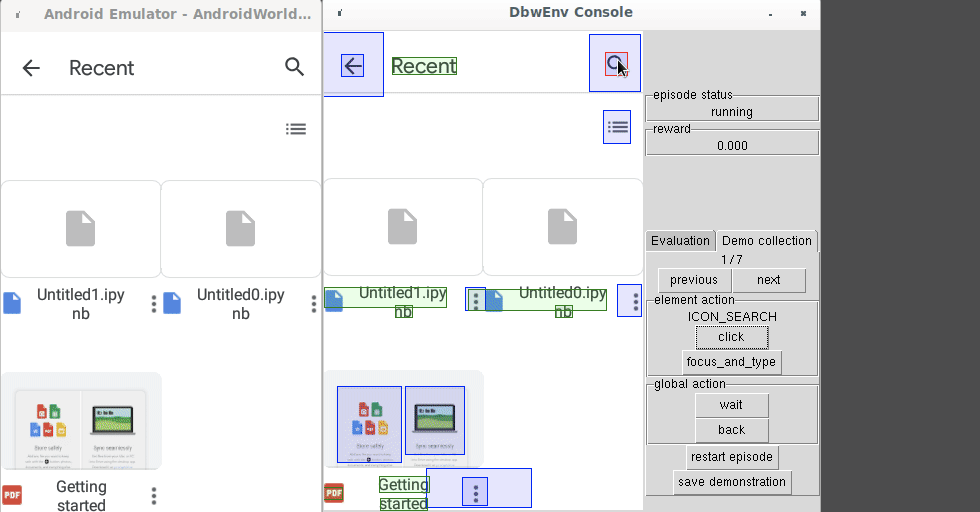}}
    \subfloat[]{\includegraphics[width=0.5\linewidth]{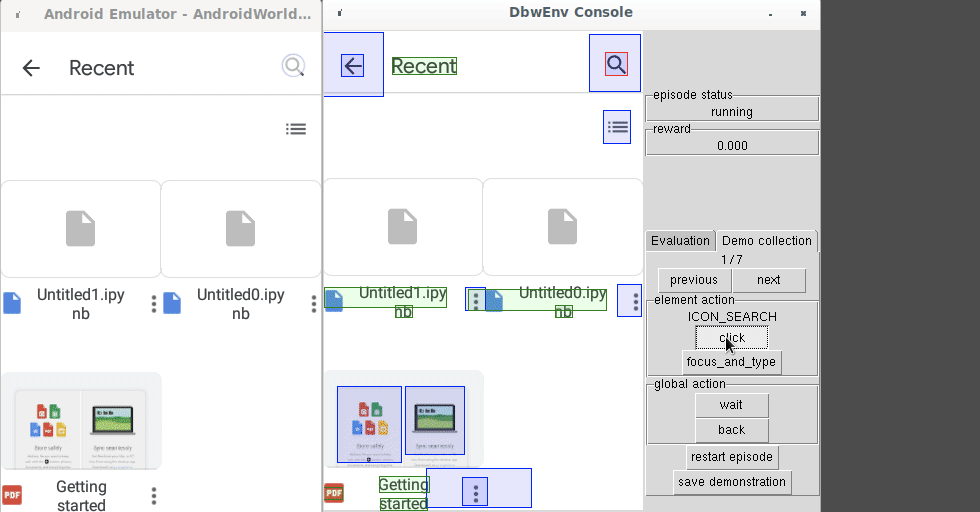}}\\
    \subfloat[]{\includegraphics[width=0.5\linewidth]{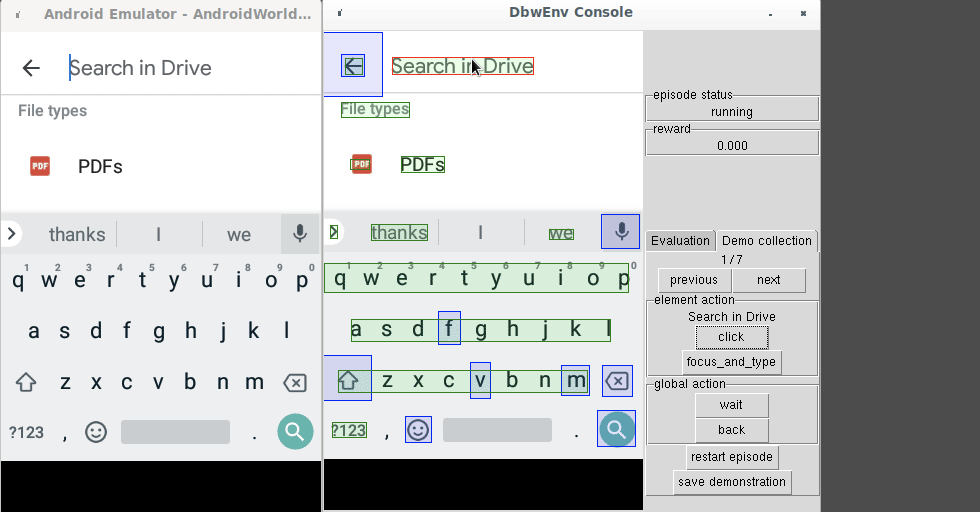}}
    \subfloat[]{\includegraphics[width=0.5\linewidth]{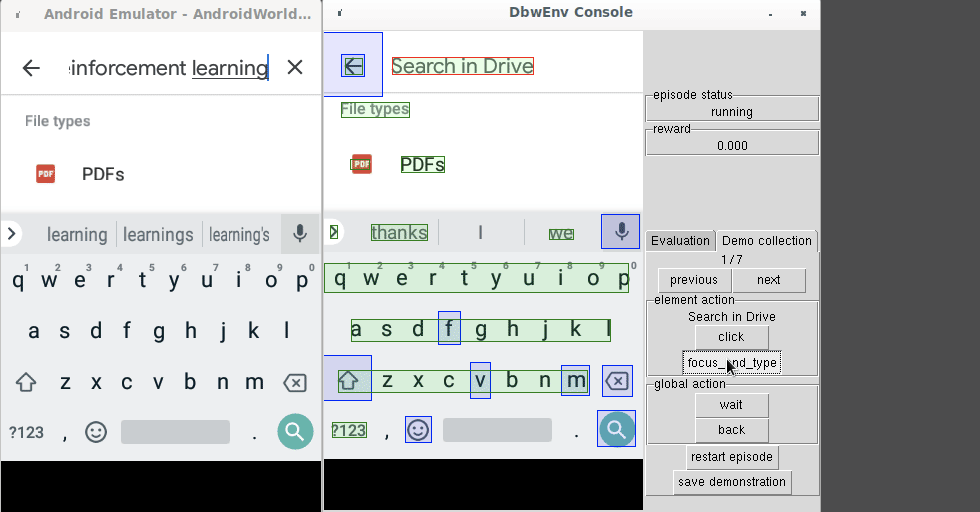}}
    \caption{Demo collection using macro actions. (a) and (b) clicking "Recent". (c) and (d) clicking the search icon. (e) and (f) focus\_and\_type "Search in Drive".}
    \label{fig:demo_collection}
\end{figure}

The GUI of the demo collection is shown in Figure \ref{fig:demo_collection}. In each image, the left portion is the screen of an Android emulator. The middle portion is the copy of the screen annotated with UI elements: icons highlighted by blue boxes and texts highlighted by green boxes. Red boxes indicate selected UI elements that are the targets for the next action. Red text (not shown in the 6 images but appear in the figures of the later part of the paper) are the action types. For an element action, the text of the action type is overlaid on the element; for an global action, the text is in the middle of the screen. The right is the control panel that has buttons to trigger agent actions.

Figure \ref{fig:demo_collection} shows the process that a human performs 3 macro actions:
\begin{enumerate}
    \item Click “Recent”: select the text label “Recent” on the annotated screen and click the “click” button on the control panel.
    \item Click the search icon: select the search icon on the annotated screen and click the “click” button on the control panel.
    \item focus\_and\_type on “Search in Drive”: select the text label “Search in Drive” and click the focus\_and\_type button on the control panel.
\end{enumerate}

Note that the human relies on the macro actions during demo collection. It is possible that the human may operate differently if the emulator is manipulated directly. The advantage is clearly that the actions in human demonstrations are in the exact same action space as that of learning agents. There is no need to recognize and convert a sequence of low-level actions to a macro action, which itself is a challenging task. For UI navigation, it is possible that there are multiple solutions for a task. We want the agents to be able to complete tasks successfully with any one of the solutions. It is not necessary to require an agent to always pick the optimal solution or be consistent with any operation style.

\subsection{Demo augmentation}

Not all the UI elements have the same importance for UI navigation. Certain elements are frequently operated for given tasks, while others are irrelevant. In particular for UI navigation, elements that help navigating, such as the back button, the home button, the menu button, and the search icon for a search task, are important. Whereas the detailed information, such as the contents of an email, or the body of a news article can be ignored.
A demonstration not only teaches an agent what actions to take for given states, but also provides information on which UI elements are more important for the task. Based on this observation, we propose demo augmentation:

\begin{enumerate}
    \item Classify all UI elements into two groups: critical and irrelevant. 
    \begin{itemize}
    \item Critical UI elements: elements receiving actions in the current step, all the icons, and common UI navigation elements.
    \item Irrelevant UI elements: all others
    \end{itemize}
    \item Randomize texts associated with randomly selected irrelevant elements.
It is simply done by replacing the word embedding vector by a random vector. For each state, a subset of the irrelevant elements are randomized with a predefined probability that we arbitrarily pick as 50\%.
    \item Add random offsets to the bounding rectangles for randomly selected irrelevant elements.
The probability that the bounding rectangle of a UI element is randomized is arbitrarily picked as 80\%.

\end{enumerate}

\subsection{Demos from screenshots}
Some failures are difficult to reproduce as they happen randomly with a small probability. Besides, we are dealing with real applications on real systems, and the applications remember the previous states.  Even for deterministic cases, unless all the operations of the previous states are recorded, there is no guarantee that an exact scenario can be reproduced. The contents of some apps change over time as they are fetched from servers. Fortunately, screenshots for those failures are easily available as they are saved during evaluation. Therefore, we collect demonstrations from individual screenshots to cover those rare cases.

There are a few disadvantages. Actions selected for a screenshot can’t be validated on an emulator, hence can be wrong, as the screenshot is not reproduced on the emulator. There is no transition between steps as in typical reinforcement learning trajectories. Besides no effort to reproduce, another advantage for screenshot demos is that it only needs human corrections for failed screenshots, not the whole episode.
For behavioral cloning, there is no change to accommodate screenshot demos, because the neural net predicts actions from screenshots/states. In standard DQfD, there are two types of training samples:
\begin{itemize}
\item Episodes from interaction with an RL environment that produce Q learning loss.
\item Episodes from demonstrations that incur both Q learning loss and classification loss.
\end{itemize}

We add a third type of training samples:

\begin{itemize}
\item Screenshot demonstrations that only contribute classification loss.
\end{itemize}

\section{Experimental Results}

\subsection{Evaluation tasks}
We have built two tasks on Android to collect demonstrations and evaluate trained agents. Each task is a wrapper of AndroidEnv\cite{androidenv} to provide reinforcement learning interfaces compatible with DeepMind’s $dm\_env$.

\subsubsection{AndroidInstall}
\begin{figure}
    \centering
    \subfloat[]{\includegraphics[width=0.5\linewidth]{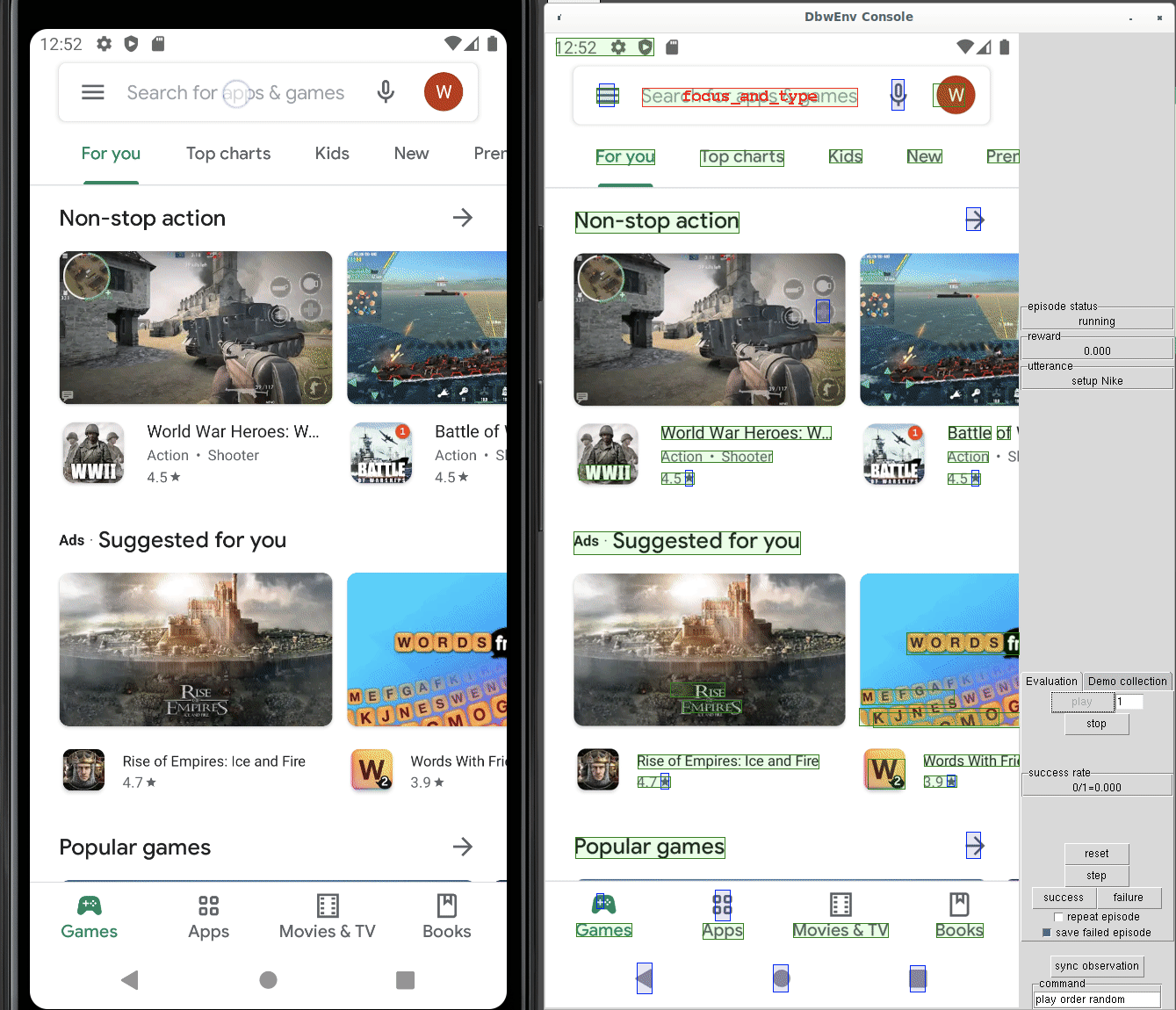}}
    \subfloat[]{\includegraphics[width=0.5\linewidth]{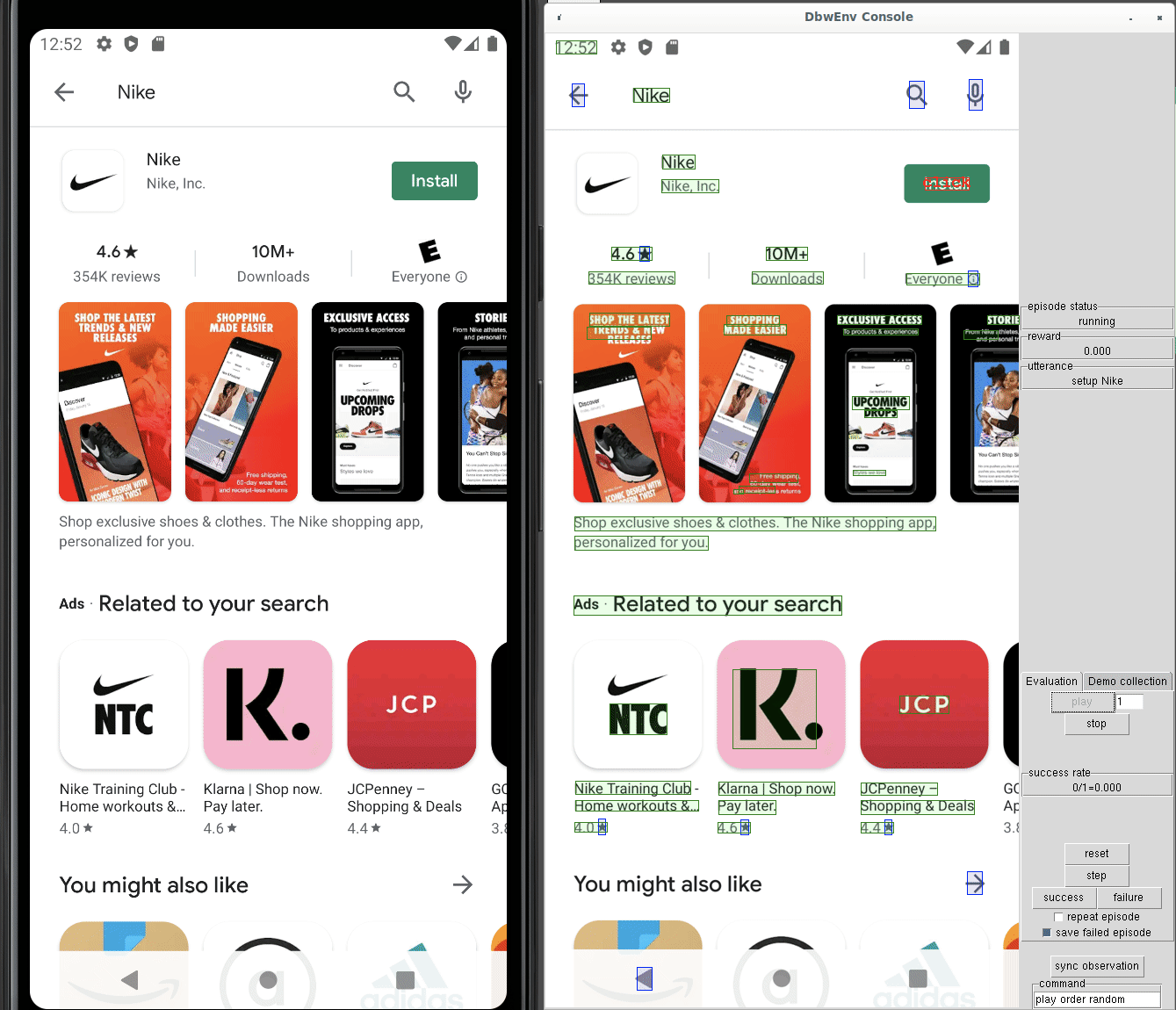}}\\
    \subfloat[]{\includegraphics[width=0.5\linewidth]{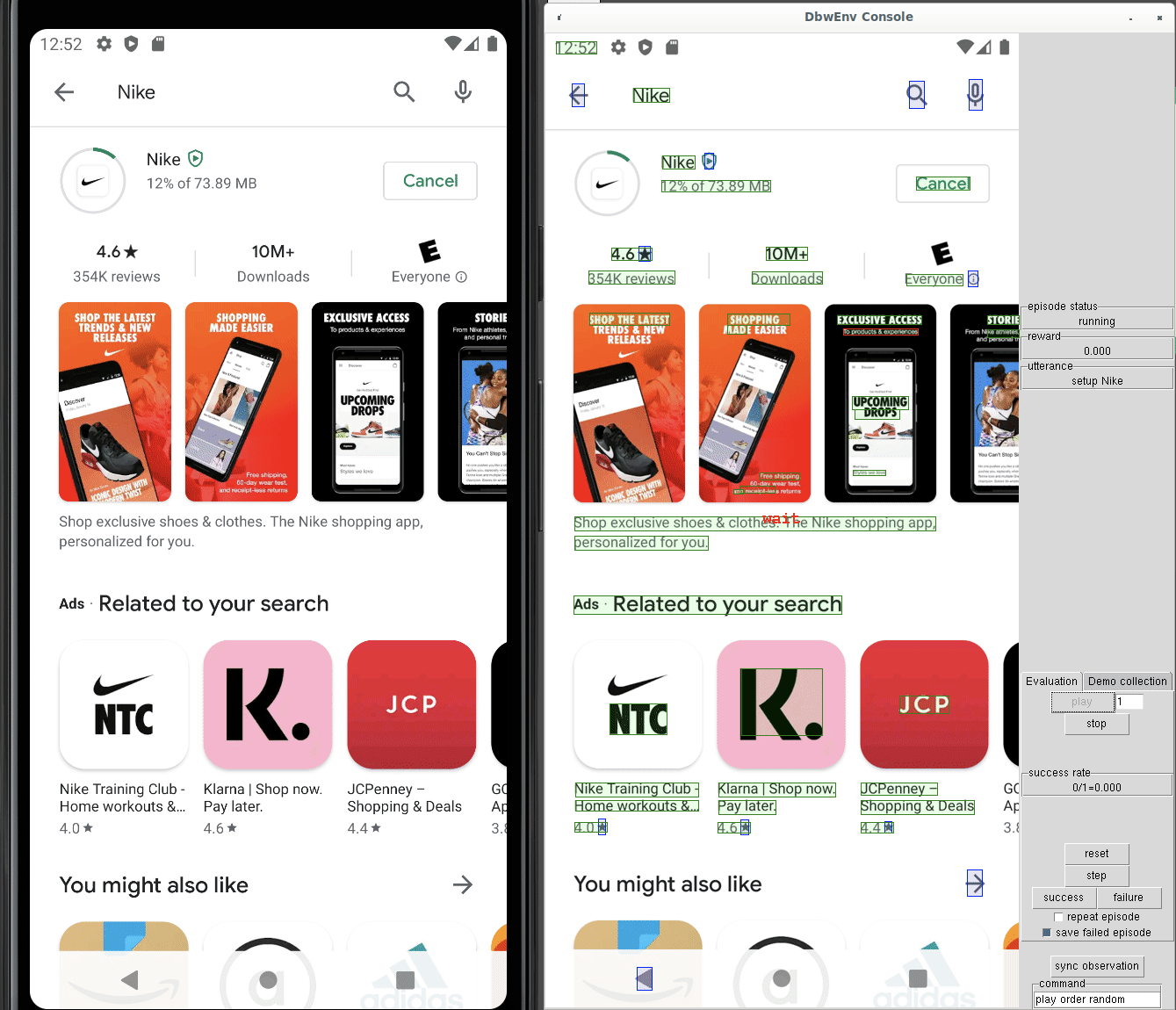}}
    \subfloat[]{\includegraphics[width=0.5\linewidth]{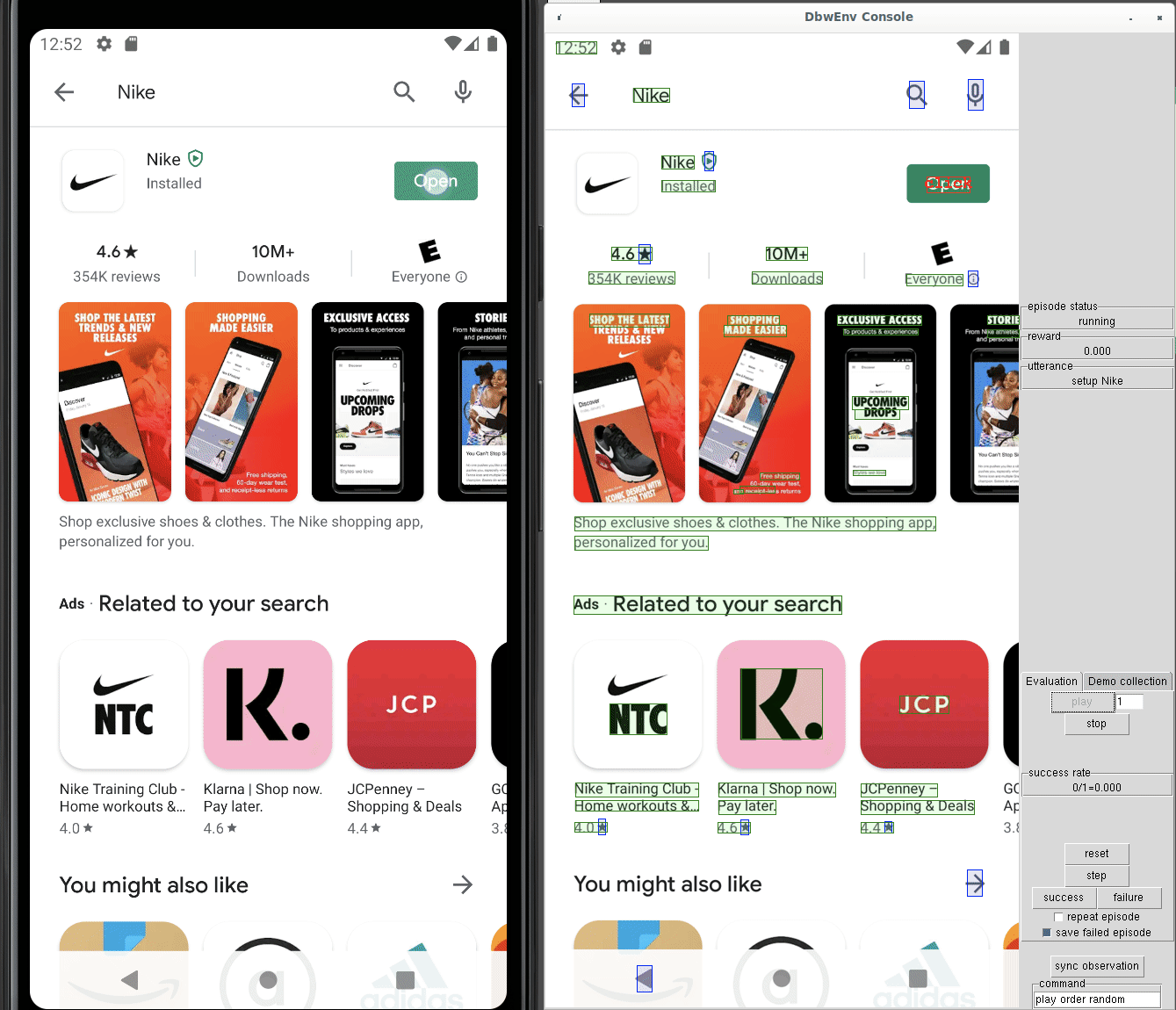}}\\
    \caption{One episode of a trained agent completing the AndroidInstall task. (a) Search app name in Google Play Store. (b) Click Enter to start installation. (c) Wait for the installation to complete. (d) Click Open to launch the app.}
    \label{fig:android_install}
\end{figure}

AndroidInstall is a task requiring an agent to install a list of Android applications using the Google Play store. At reset of each episode, Google Play Store is launched, and an utterance in the form of “install $<$application name$>$” is sent to the agent, where the application name is picked from the list, either randomly or in sequential order. The agent is expected to search the application in the play store, and to proceed from the page of the app. Once an application is installed, the agent should launch it. If  the current package name matches the expected value, the episode is declared as successful. The image sequence in Figure \ref{fig:android_install} shows an episode of the installation.

AndroidInstall has two goals: 1) providing a consistent metric for evaluating architecture and algorithm changes. 2) setting up the environment by installing all the needed apps for other tasks, such as AndroidSearch. Therefore, we deliberately reduce the diversity of the task. Viewing parameters stay constant across episodes. The play store app is force closed and restarted for each episode to simplify initial states of each episode.

\subsubsection{AndroidSearch}

In this task, a randomly selected application or web site from a pool of about 80 is launched at the reset of each episode with randomized viewing parameters, including pixel density, font scale, and orientation. A sequence of clicks are then applied to randomly selected UI elements to further randomize the starting state of an episode. An utterance in the form of “search for x” is passed to the agent, where x is randomly picked and may depend on the current application. An agent is expected to input the search phrase into the nearest search bar. If there is no search bar on the current screen, the agent should first reveal it. Once the required search phrase is entered successfully and either relevant results appear on the screen or a message indicating nothing is found, the episode is declared as successful by the task environment. Note that all the apps and websites are real with dynamic contents and their GUI may change due to updates. This is more challenging than environments whose variations are fixed.

\subsection{Agent performance}

\newif\ifshow 
\showfalse 

\ifshow
  \includecomment{wrap}
\else
  \excludecomment{wrap} 
\fi

\begin{wrap}
Figure \ref{fig:success_rate_vs_number_of_demos} shows the relationship between success rate and the number of demonstrations used for training agents on  AndroidSearch. Unsurprisingly, more demonstrations generally result in better performance. More importantly, the success rate goes beyond 97\% with just 1500 demonstration episodes, which is orders of magnitude fewer than typical RL setup. We believe that a critical factor is the simplified state space and action space in our design. The first half of demonstrations are collected from 16 apps, while the second half uses all the 100 or so apps. The agent has certain capabilities of generalizing to unseen apps, as the success rate of 37\%, which is obtained by training with demonstrations from 16 apps, is significantly larger than $16 / 100 = 16\%$. However, we observe that the workflows in different apps have very large variance. Therefore for practical purposes, all the distinctive patterns should be included in the training samples. Fortunately, our agent only requires a small amount of demonstrations that can be collected in hours from scratch.

\begin{figure}
\centering
\includegraphics[width=0.9\columnwidth]{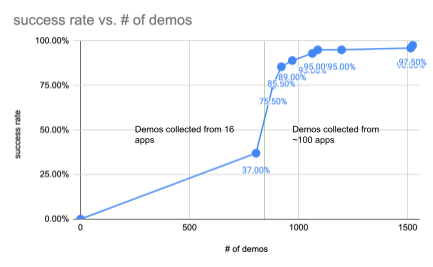} 
\caption{Success rates of the search agent vs. the number of demonstrations.}.
\label{fig:success_rate_vs_number_of_demos}
\end{figure}
\end{wrap}

Table \ref{tbl:demo_augmentation_install} shows the effect of demo augmentation on AndroidInstall. In this setup, only 16 human demo episodes are collected. Demo augmentation increases the total number of demo episodes to 1600. Clearly both DQfD and BC reach significantly higher success rates with demo augmentation.

\begin{table}[!htbp]
\centering
\begin{tabular}{*5c}
\toprule
  &  \multicolumn{2}{c}{Augmented} & \multicolumn{2}{c}{Without augmentation}\\
\midrule
Agent          & DQfD   & BC    & DQfD   & BC\\
Success rate   &  99.0\% & 96.6\%   & 80.9\%  & 89.5\%\\
\bottomrule
\end{tabular}
\caption{Success rates of agents trained from 16 human demos on AndroidInstall.}
\label{tbl:demo_augmentation_install}
\end{table}

We actually started with just two demo episodes to train an AndroidInstall agent, one from a fresh installation of an app, and one from updating an existing installation. With augmentation, the success rate reached 90\% by using just the two demo episodes. To push up the success rate, more demo episodes are added to deal with various corner cases, such as pop up ads and system messages. When the number of demo episodes reaches 16, the success rate is near perfect.

\begin{wrap}
Similarly, demo augmentation is also helpful for AndroidSearch. As shown in Figure \ref{fig:demo_augmentation_search}, the success rates of a DQfD agent are 92\% with augmentation and 85\% without when trained on 117 demos. The percentages become 97\% vs 91.5\% when the agents are trained on 220 demos. 

\begin{figure}
\centering
\includegraphics[width=0.9\columnwidth]{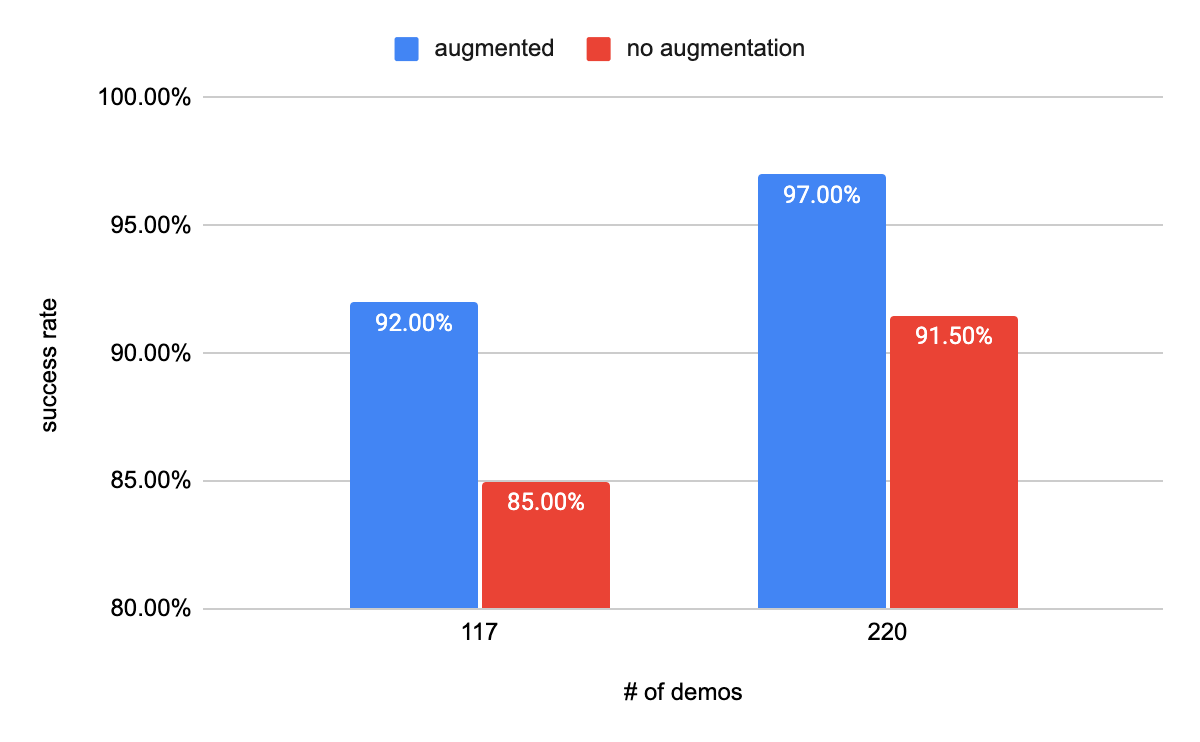} 
\caption{Demo augmentation on the agents of the search task.}.
\label{fig:demo_augmentation_search}
\end{figure}
\end{wrap}

Table \ref{tbl:training_options} lists the success rates of DQfD agents trained using different options on the search task. The options include:
\begin{itemize}
    \item Demo augmentation: whether demo augmentation is used;
    \item Screenshot: whether screenshot demos are included in the training samples
    \item Action mask: whether the loss function considers action mask as shown in formula \ref{eq:1} 
\end{itemize}
. The training samples are composed of 425 full episodes and 261 screenshots. When all the three options are enabled, the success rate reaches 98.7\%.

\begin{table}
\centering
\begin{tabular}{l|l|l|l}
\toprule
Demo & Screenshot & Action mask & Success\\
augmentation &      &      & rate \\
\midrule
No & No & No & 94.9\% \\
No & No & Yes & 96.2\% \\
Yes & No & Yes & 97.5\% \\
Yes & Yes & Yes & 98.7\% \\
\bottomrule
\end{tabular}
\caption{Success rates of DQfD on AndroidSearch using different training options.}
\label{tbl:training_options}
\end{table}

Tables \ref{table2} and \ref{table3} compare the success rates of the two configurations: fewer demos with augmentation and more demos without augmentation. For AndroidInstall, the agent trained with fewer but augmented demos performs notably better than the agent trained with 30x more demos but without augmentation (99\% vs 95.6\%). For AndroidInstall, the agent trained with 220 demos with augmentation matches the success rate of the agent trained with 1500 demos without augmentation.

\begin{table}
\centering
\begin{tabular}{l|l}
\toprule
augmented from 16 & 512 human demos\\
human demos &  without augmentation \\
\midrule
99\% & 95.6\%\\
\bottomrule    
\end{tabular}
\caption{Success rate comparison of DQfD on AndroidInstall: fewer demos with augmentation vs. more demos without augmentation.}
\label{table2}
\end{table}

\begin{table}
\centering
\begin{tabular}{l|l}
\toprule
augmented from  & 1500 human demos \\
220 human demos &  without augmentation\\
\midrule
97\% & 97.5\%\\
\bottomrule
\end{tabular}
\caption{Success rates comparison of DQfD on AndroidSearch: fewer demos with augmentation vs. more demos without augmentation.}
\label{table3}
\end{table}

\subsection{Selected cases of a DQfD agent doing searches}

\begin{figure*}
    \centering
    \subfloat[]{\includegraphics[width=0.198\linewidth]{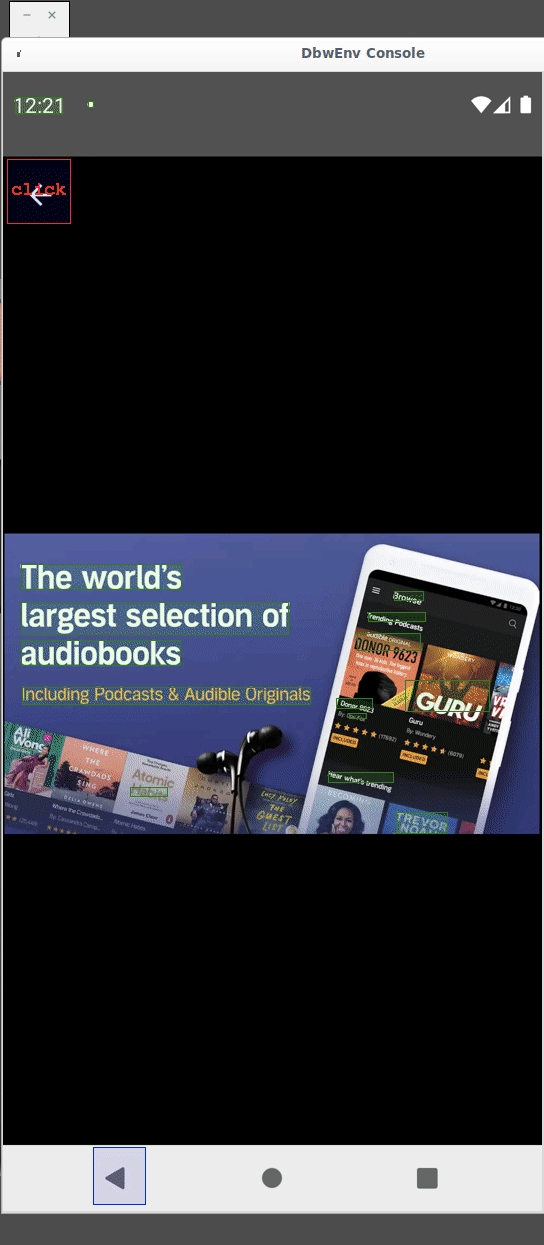}}
    \subfloat[]{\includegraphics[width=0.198\linewidth]{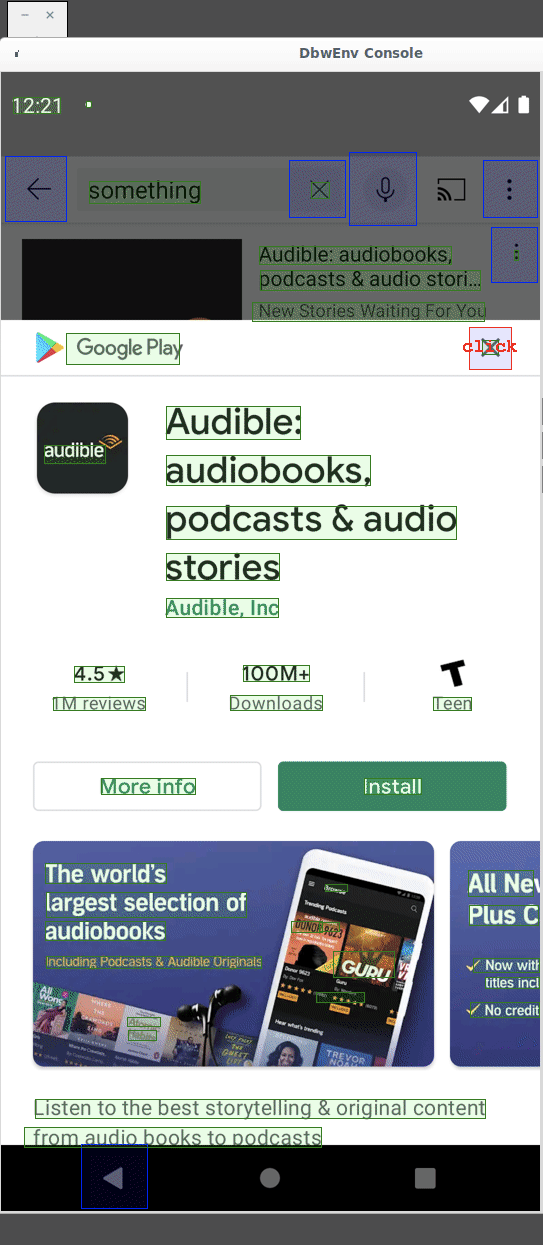}}
    \subfloat[]{\includegraphics[width=0.2\linewidth]{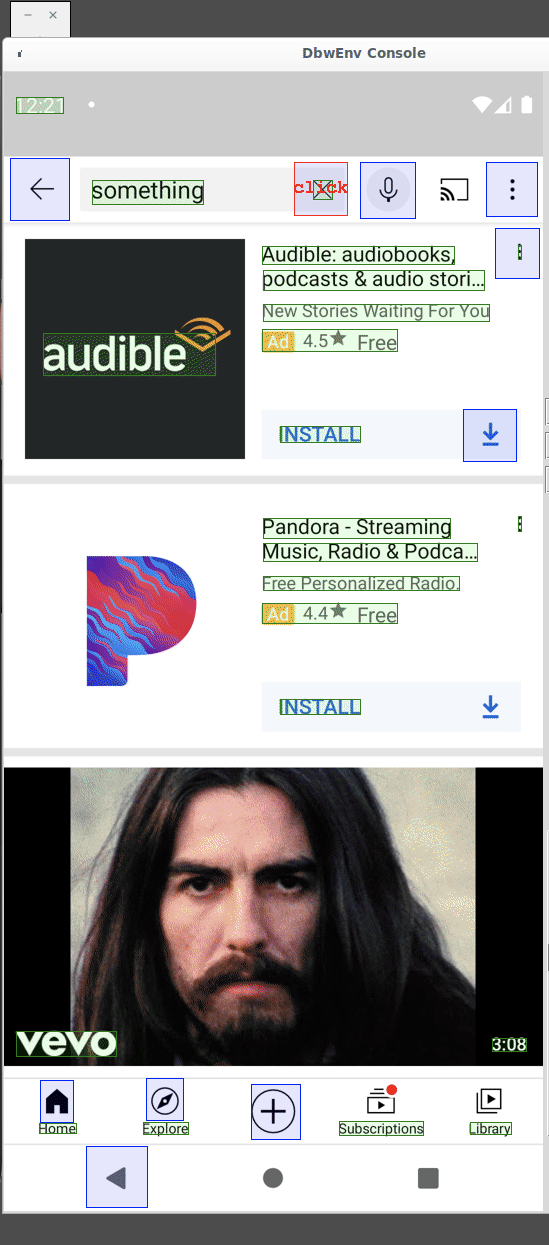}}
    \subfloat[]{\includegraphics[width=0.195\linewidth]{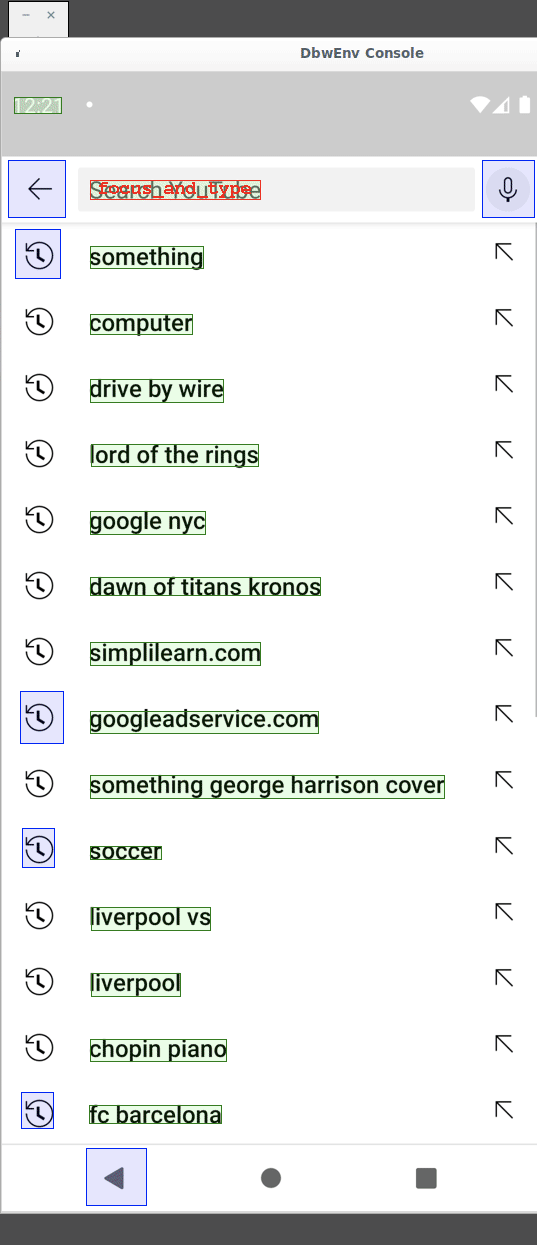}}
    \subfloat[]{\includegraphics[width=0.212\linewidth]{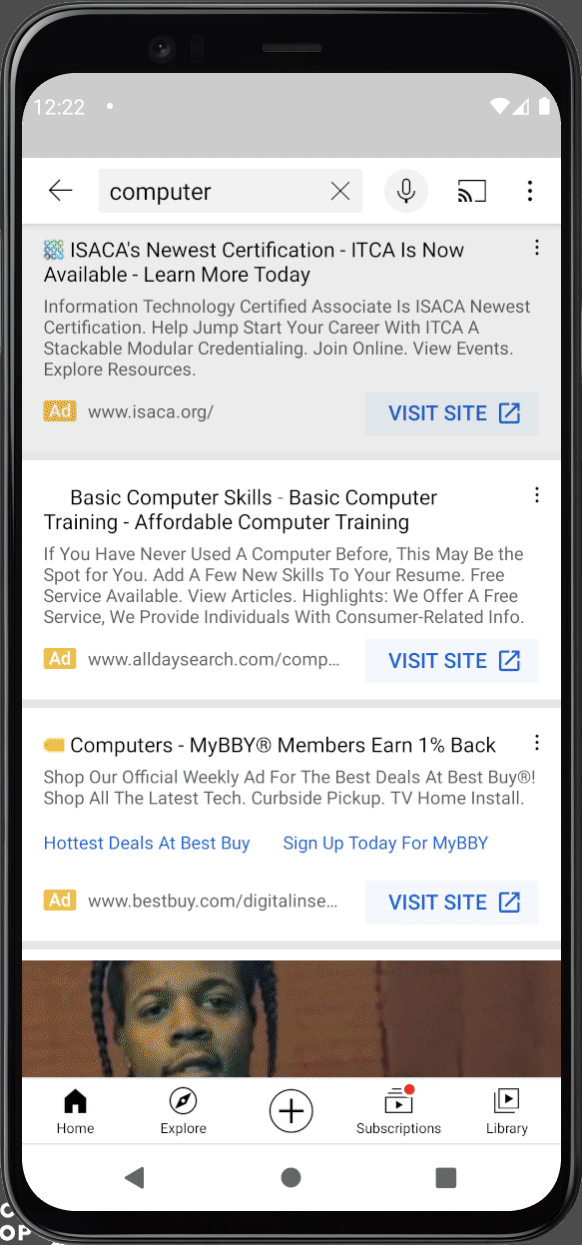}}
    \caption{Agent searches in YouTube. (a) Click the back button to dismiss a popup ads. (b) Click "X" to dismiss the install page of Audible. (c) Click "X" to erase the previously entered search phrase "something". (d) and (e) Enter the new search phrase "computer" and wait for the search results.}
    \label{fig:search_youtube}
\end{figure*}

\begin{figure*}
    \centering
    \subfloat[]{\includegraphics[width=0.201\linewidth]{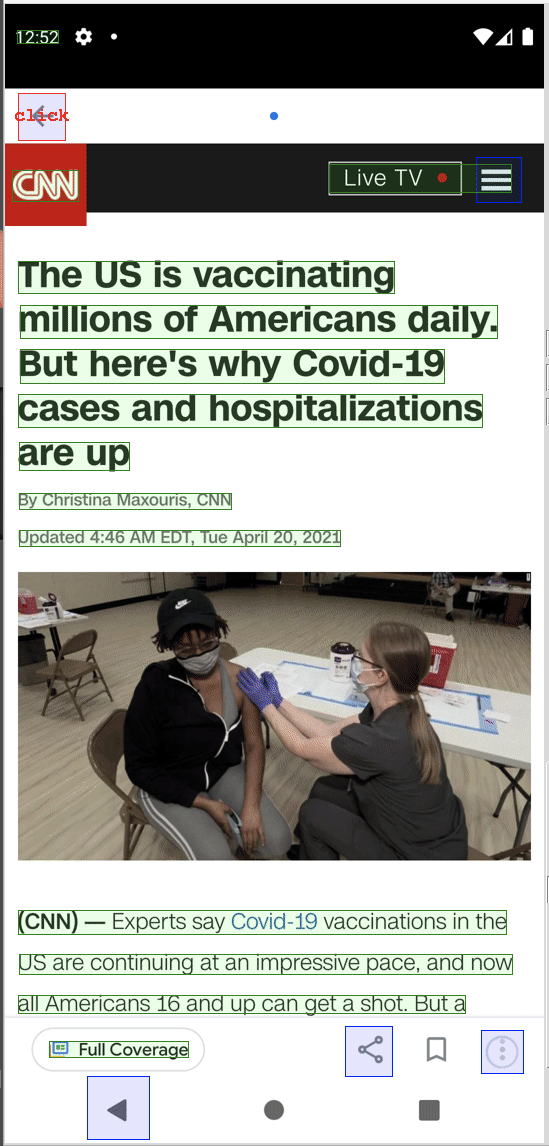}}
    \subfloat[]{\includegraphics[width=0.2\linewidth]{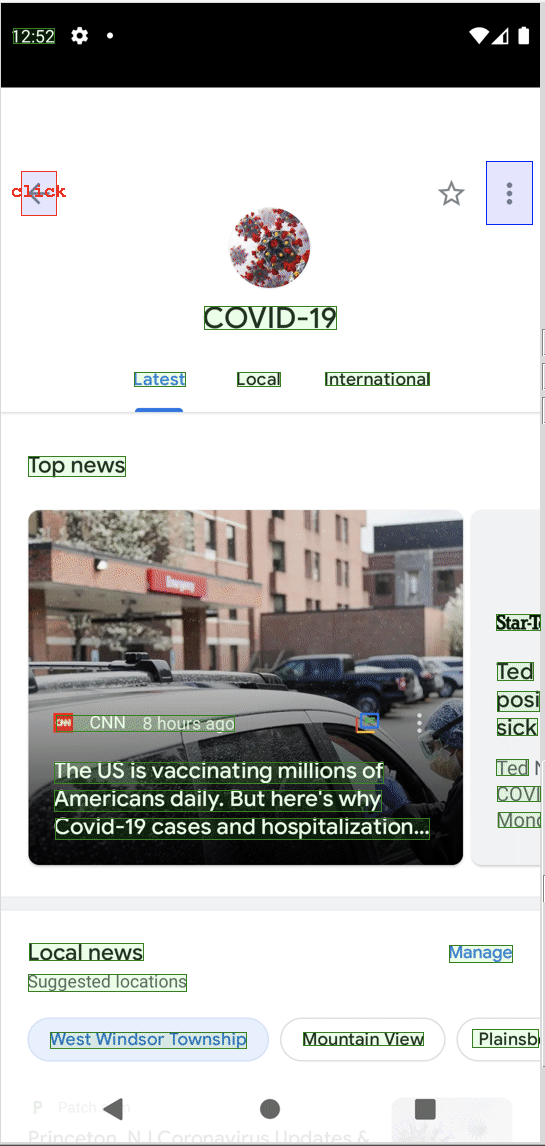}}
    \subfloat[]{\includegraphics[width=0.2\linewidth]{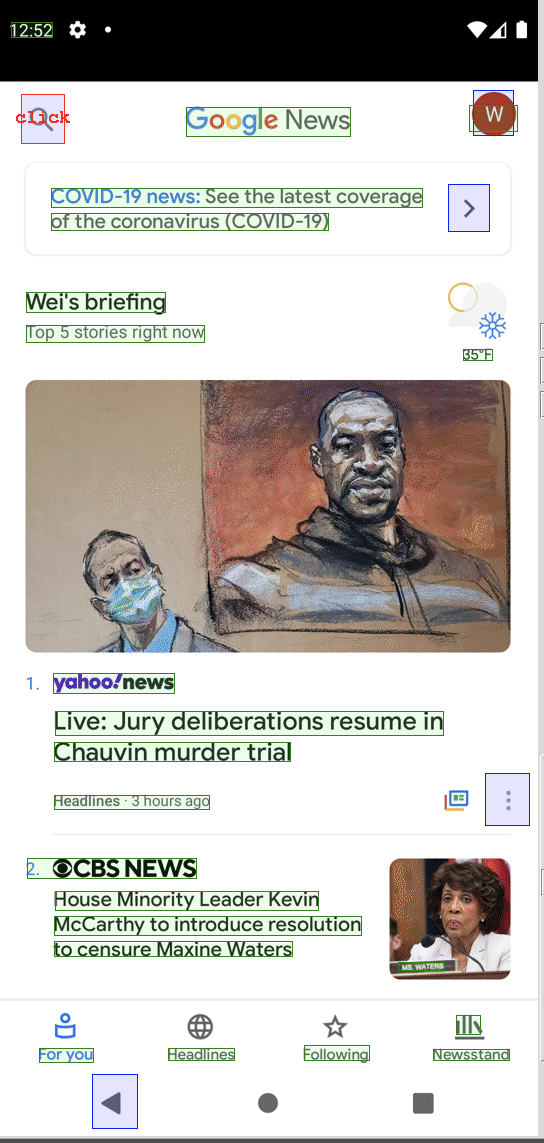}}
    \subfloat[]{\includegraphics[width=0.2\linewidth]{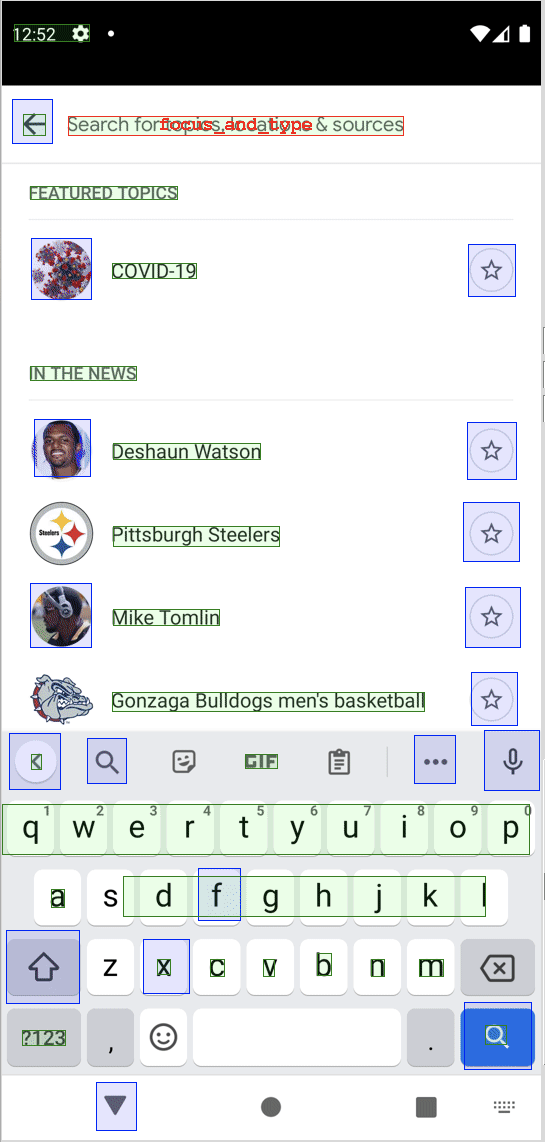}}
    \subfloat[]{\includegraphics[width=0.198\linewidth]{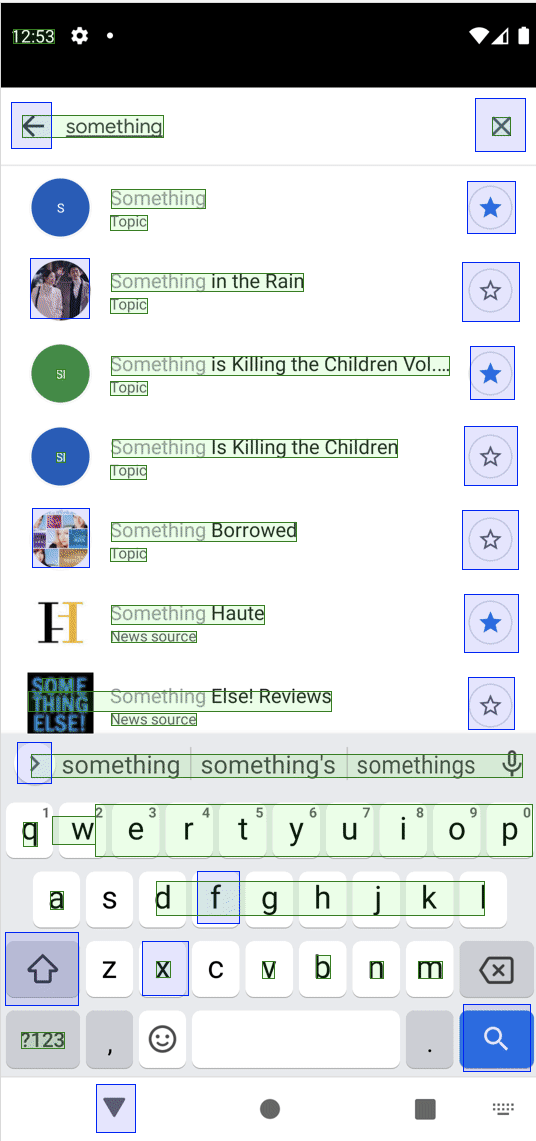}}
    \caption{Agent searches in Google News. (a) and (b) Click the back button twice to go to the home screen where the search icon is available. (c) Click the search icon to reveal the search bar. (d) Enter search phrase in the search bar. (e) Wait for the search results.}
    \label{fig:search_news}
\end{figure*}

Figure \ref{fig:search_youtube} and \ref{fig:search_news} show a few examples of a trained DQfD agent performing searches. Note that the red boxes and texts indicate the agent choices of UI elements to operate and the action types.

Figure \ref{fig:search_youtube} shows an episode in YouTube. The agent dismisses popups twice to reveal the search bar (a) and (b). It then clicks the X button to erase the previous search phrase “something” (c). Next, it sets focus and enters the required search phrase “computer” (d). Finally, the results of the search is loaded (e).

Figure \ref{fig:search_news} shows a search episode in Google News. The agent clicks the back arrow twice to go to the main screen (a) and (b). Then it clicks the search icon to bring out the search bar (c). Next, it sets focus to the search bar (d) and enters the search phrase “something” (e).

\begin{wrap}
Figure \ref{fig:search_facebook} shows an episode in the Facebook app, the agent clicks the back arrow to go to the previous screen. Then it uses the X button to erase the previous search phrase “computer”. Next it focuses on the search bar and enters the new search phrase “something”.

\begin{figure*}
    \centering
    \subfloat[]{\includegraphics[width=0.201\linewidth]{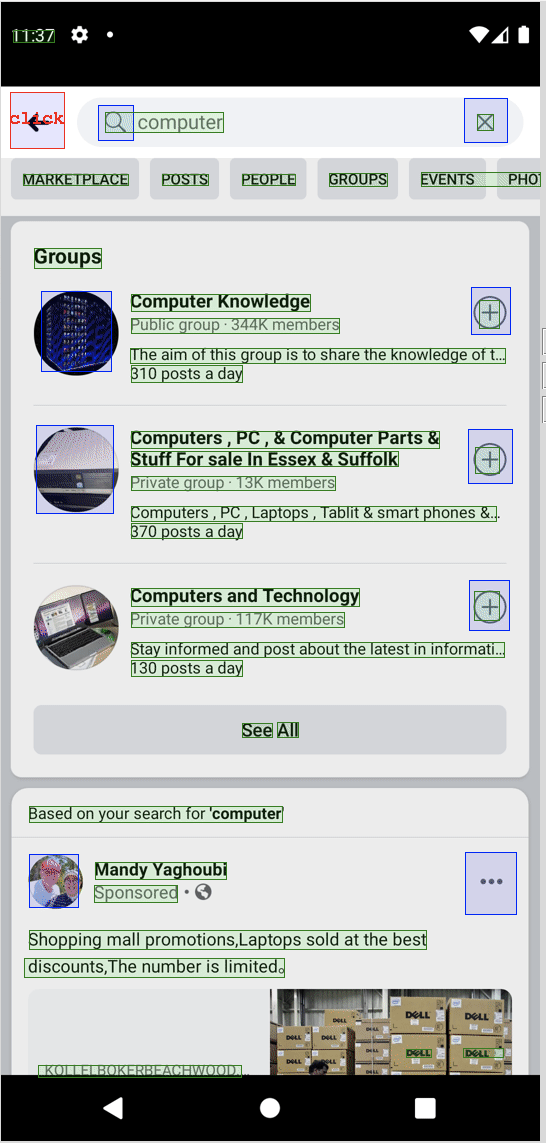}}
    \subfloat[]{\includegraphics[width=0.2\linewidth]{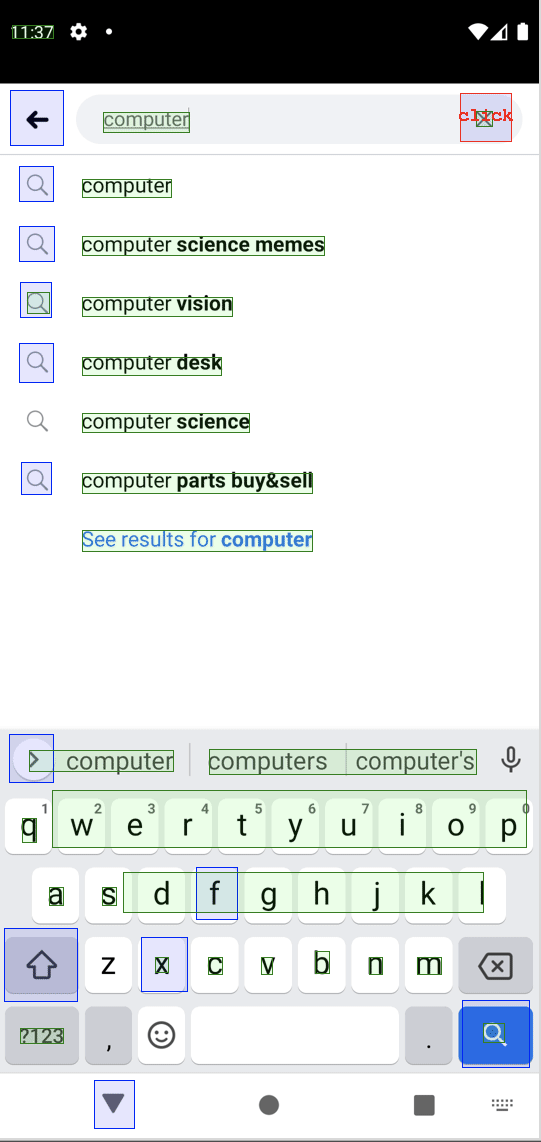}}
    \subfloat[]{\includegraphics[width=0.2\linewidth]{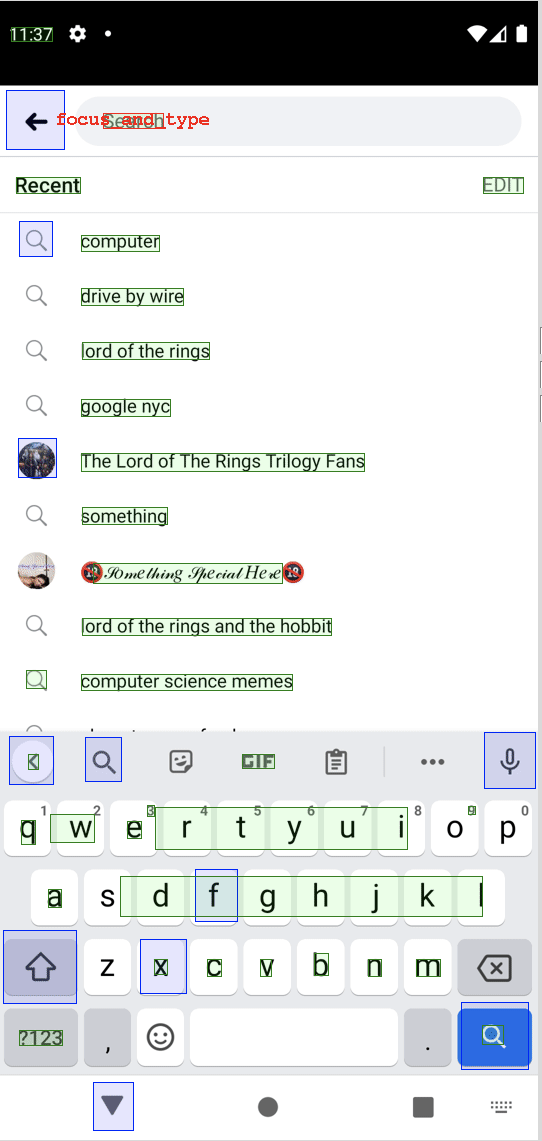}}
    \subfloat[]{\includegraphics[width=0.2\linewidth]{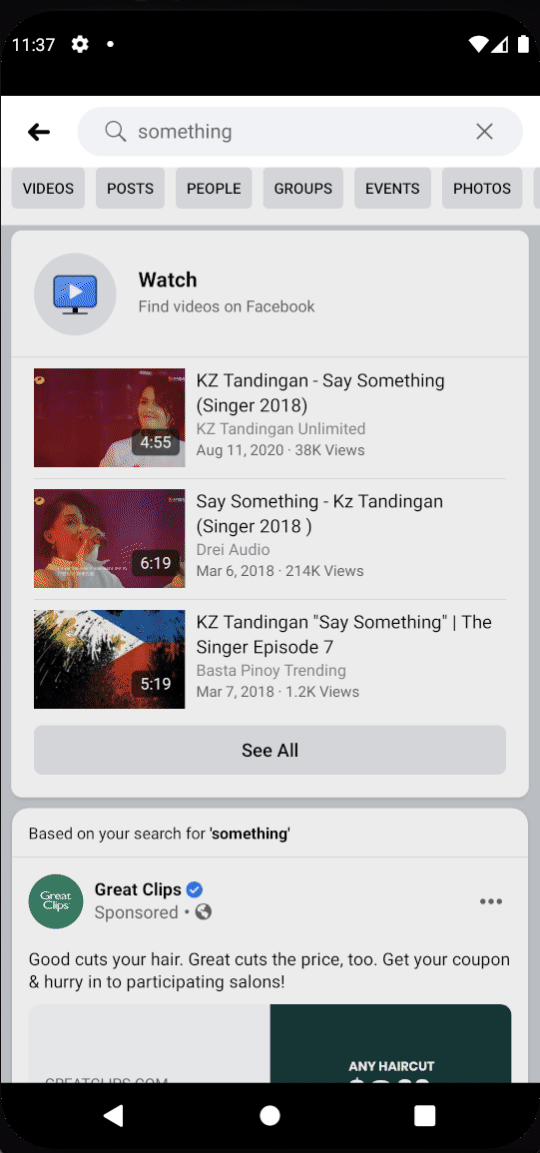}}
    \caption{Agent searches in Facebook.}
    \label{fig:search_facebook}
\end{figure*}
\end{wrap}
\section{Conclusion and future work}

In this paper, we present an end-to-end workflow of building UI navigation agents from demonstrations, including definition of state and action space, neural network architecture, macro actions, error-driven demo collection, and demo augmentation. Our experimental results prove that the agents reach high success rates by training from a small number of demonstration episodes.

Our goal is to develop learning-based techniques to replace existing UI automation macros handcrafted by software engineers. Our next step will shift more towards engineering by experimenting our agent building technique on more automation tasks that have practical purposes. We plan to start with training individual agents independently for different tasks. But we will certainly research in the possibilities of multi-task agents that are conditioned on the utterances where mapping natural language instructions to sub-tasks become critical. One expected challenge is how to maintain the high success rate and ease of training while making the agents generalize better.

It is not uncommon that a correct action fails on a real system, such as Android emulator, due to occasional system unresponsiveness. In such a case, it is desirable that an agent retries a failed action, sometimes with multiple attempts. A memory-less agent naturally repeats the same action for the same state. Besides to simplify neural net, this is another reason that we deliberately avoid memory mechanism, such as LSTM. The drawback is that an agent may be trapped in repeating the same wrong action that has no effect. In the future, we will research techniques that savages agents without significantly increasing the neural net complexity. One thought is to feed the count of actions that are performed for the current episode, and encourage agent to explore alternatives during inference.

\bibliographystyle{unsrtnat}
\bibliography{uinav}  

\begin{thebibliography}{11}
\providecommand{\natexlab}[1]{#1}

\bibitem[{Branavan et~al.(2009)Branavan, Chen, Zettlemoyer, and
  Barzilay}]{mapping_instruction_to_action}
Branavan, S.; Chen, H.; Zettlemoyer, L.~S.; and Barzilay, R. 2009.
\newblock Reinforcement Learning for Mapping Instructions to Actions.
\newblock \emph{Proceedings of the Joint Conference of the 47th Annual Meeting
  of the ACL and the 4th International Joint Conference on Natural Language
  Processing of the AFNLP}.

\bibitem[{Branavan, Zettlemoyer, and
  Barzilay(2010)}]{reading_between_the_lines}
Branavan, S.; Zettlemoyer, L.; and Barzilay, R. 2010.
\newblock Reading between the Lines: Learning to Map High-Level Instructions to
  Commands.
\newblock In \emph{Proceedings of the 48th Annual Meeting of the Association
  for Computational Linguistics}, 1268--1277. Uppsala, Sweden: Association for
  Computational Linguistics.

\bibitem[{Gur et~al.(2021)Gur, Jaques, Malta, Tiwari, Lee, and
  Faust}]{gminiwob}
Gur, I.; Jaques, N.; Malta, K.; Tiwari, M.; Lee, H.; and Faust, A. 2021.
\newblock Adversarial Environment Generation for Learning to Navigate the Web.
\newblock \emph{CoRR}, abs/2103.01991.

\bibitem[{Gur et~al.(2018)Gur, R{\"{u}}ckert, Faust, and
  Hakkani{-}T{\"{u}}r}]{learning_web_nav}
Gur, I.; R{\"{u}}ckert, U.; Faust, A.; and Hakkani{-}T{\"{u}}r, D. 2018.
\newblock Learning to Navigate the Web.
\newblock \emph{CoRR}, abs/1812.09195.

\bibitem[{Hester et~al.(2017)Hester, Vecer{\'{\i}}k, Pietquin, Lanctot, Schaul,
  Piot, Sendonaris, Dulac{-}Arnold, Osband, Agapiou, Leibo, and Gruslys}]{dqfd}
Hester, T.; Vecer{\'{\i}}k, M.; Pietquin, O.; Lanctot, M.; Schaul, T.; Piot,
  B.; Sendonaris, A.; Dulac{-}Arnold, G.; Osband, I.; Agapiou, J.~P.; Leibo,
  J.~Z.; and Gruslys, A. 2017.
\newblock Learning from Demonstrations for Real World Reinforcement Learning.
\newblock \emph{CoRR}, abs/1704.03732.

\bibitem[{Li et~al.(2020)Li, He, Zhou, Zhang, and Baldridge}]{nl_to_ui_action}
Li, Y.; He, J.; Zhou, X.; Zhang, Y.; and Baldridge, J. 2020.
\newblock Mapping Natural Language Instructions to Mobile {UI} Action
  Sequences.
\newblock \emph{CoRR}, abs/2005.03776.

\bibitem[{Liu et~al.(2018)Liu, Guu, Pasupat, Shi, and Liang}]{wge}
Liu, E.~Z.; Guu, K.; Pasupat, P.; Shi, T.; and Liang, P. 2018.
\newblock Reinforcement Learning on Web Interfaces Using Workflow-Guided
  Exploration.
\newblock \emph{CoRR}, abs/1802.08802.

\bibitem[{Osa et~al.(2018)Osa, Pajarinen, Neumann, Bagnell, Abbeel, and
  Peters}]{osa:2018}
Osa, T.; Pajarinen, J.; Neumann, G.; Bagnell, J.~A.; Abbeel, P.; and Peters, J.
  2018.
\newblock An Algorithmic Perspective on Imitation Learning.
\newblock \emph{Foundations and Trends in Robotics}, 7(1-2): 1–179.

\bibitem[{Pomerleau(1989)}]{pomerleau:alvinn}
Pomerleau, D.~A. 1989.
\newblock {ALVINN:} An Autonomous Land Vehicle in a Neural Network.
\newblock In Touretzky, D.~S., ed., \emph{Advances in Neural Information
  Processing Systems 1}, 305--313. San Francisco, CA: Morgan Kaufmann.

\bibitem[{Shi et~al.(2017)Shi, Karpathy, Fan, Hernandez, and Liang}]{miniwob}
Shi, T.; Karpathy, A.; Fan, L.; Hernandez, J.; and Liang, P. 2017.
\newblock World of Bits: An Open-Domain Platform for Web-Based Agents.
\newblock In Precup, D.; and Teh, Y.~W., eds., \emph{Proceedings of the 34th
  International Conference on Machine Learning}, volume~70 of \emph{Proceedings
  of Machine Learning Research}, 3135--3144. PMLR.

\bibitem[{Toyama et~al.(2021)Toyama, Hamel, Gergely, Comanici, Glaese, Ahmed,
  Jackson, Mourad, and Precup}]{androidenv}
Toyama, D.; Hamel, P.; Gergely, A.; Comanici, G.; Glaese, A.; Ahmed, Z.;
  Jackson, T.; Mourad, S.; and Precup, D. 2021.
\newblock AndroidEnv: {A} Reinforcement Learning Platform for Android.
\newblock \emph{CoRR}, abs/2105.13231.

\end{thebibliography}






\section*{Acknowledgements}
We would thank Daniel Rodriguez, Chris Rawles, and Alice Li for periodical discussions and proof reading of the write up. Our work has critical dependency on AndroidEnv and is impossible without the help from Daniel Toyama, Philippe Hamel, and Anita Gergely. We also greatly appreciate the valuable comments and suggestions from many colleagues of Google and DeepMind, especially (in alphabetic order of last names) Zafarali Ahmed, Jindong Chen, Aleksandra Faust, Mia Glaese, Izzeddin Gur, Timothy Lillicrap, James Stout, Gabriel Taubman, Manoj Tiwari, Gregory Wayne, and Xiaoxue Zang.

\end{document}